\pdfoutput=1

\documentclass[11pt]{article}

\usepackage{acl}

\usepackage{times}
\usepackage{latexsym}

\usepackage{soul}
\usepackage{url}
\usepackage{graphicx}
\usepackage{amsmath}
\usepackage{amsthm}
\usepackage{booktabs}
\usepackage{algorithm}
\usepackage{algorithmic}
\usepackage[switch]{lineno}

\usepackage{amsfonts}
\usepackage{textcomp}
\usepackage{xcolor}
\usepackage{multirow,booktabs, hhline}

\usepackage[T1]{fontenc}

\usepackage[utf8]{inputenc}

\usepackage{microtype}

\title{DeRisk: An Effective Deep Learning Framework for Credit Risk Prediction over Real-World Financial Data}

\author{Yancheng Liang$^{13}$\thanks{Both authors contributed equally to this research.} \quad Jiajie Zhang$^1{}^*$ \quad Hui Li$^2$ \quad Xiaochen Liu$^2$ \\ \textbf{Yi Hu}$^2$ \quad \textbf{Yong Wu}$^2$ \quad \textbf{Jinyao Zhang}$^2$ \quad \textbf{Yongyan Liu}$^2$ \quad \textbf{Yi Wu}$^{13}$ \\
  $^1$ Tsinghua University \quad $^2$ Fintopia Group \quad $^3$ Shanghai Qi Zhi Institute
}

\begin{document}
\maketitle
\begin{abstract}
Despite the tremendous advances achieved over the past years by deep learning techniques, the latest risk prediction models for industrial applications still rely on highly hand-tuned stage-wised statistical learning tools, such as gradient boosting and random forest methods. Different from images or languages, real-world financial data are high-dimensional, sparse, noisy and extremely imbalanced, which makes deep neural network models particularly challenging to train and fragile in practice. In this work, we propose \emph{DeRisk}, an effective deep learning risk prediction framework for credit risk prediction on real-world financial data. \emph{DeRisk} is the first deep risk prediction model that outperforms statistical learning approaches deployed in our company's production system. We also perform extensive ablation studies on our method to present the most critical factors for the empirical success of \emph{DeRisk}.
\end{abstract}

\section{Introduction}

\begin{figure*}[!t]
  \centering
  \includegraphics[width=\textwidth]{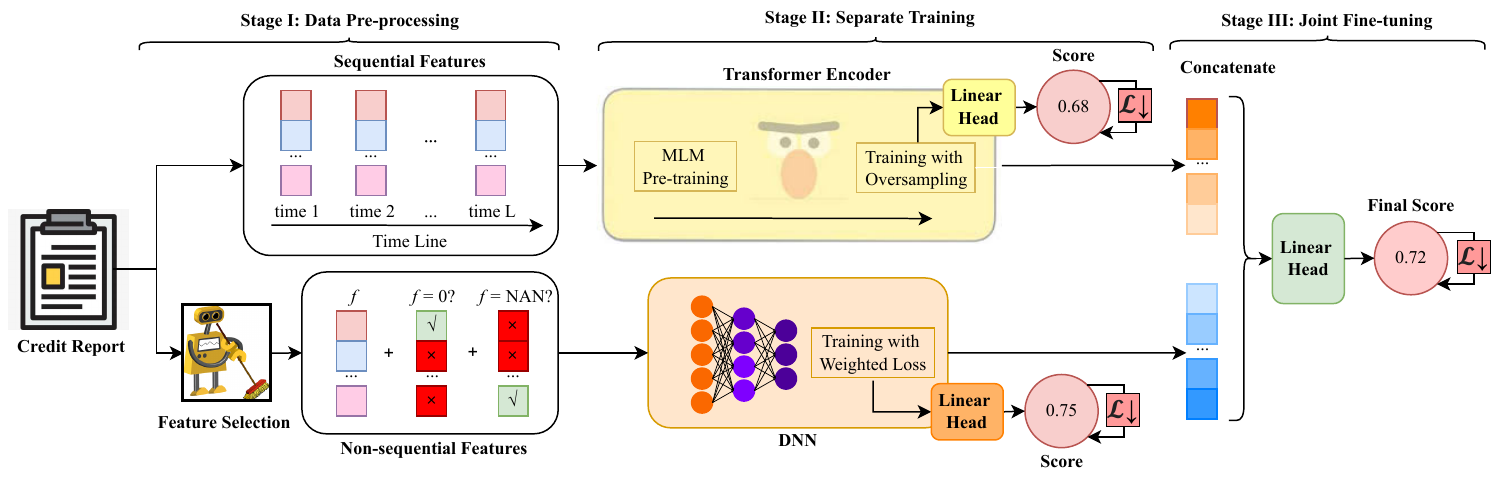}
  \vspace{-2mm}
  \caption{The three-stage pipeline of our \emph{DeRisk} framework. First, in data pre-processing, feature selection and DL-specific data argumentation are adopted to benefit the optimization of DL models. Then we separately train two models for non-sequential data and sequential data, respectively. Finally, we combine them and fine-tune the joint model on the whole multi-format data. 
  In contrast to the end-to-end paradigm in conventional DL applications, we remark the multi-stage process is critical to the overall success of \emph{DeRisk} on real-world financial data.}
  \vspace{-1mm}
  \label{fig:pipline}
\end{figure*}

Credit risk is the risk of loan default or loan delinquency when a borrower fails to repay on time. Credit risk prediction is an analytical problem that is vital for financial institutions when they are formulating lending strategies for loan applications. It helps make lending decisions by assessing the solvency of the applicants from their credit information. Accurate prediction keeps bad debts at a low level, which directly saves substantial financial loss for the multi-billion dollar credit loan industry \cite{malekipirbazari2015risk,tan2018deep}. As credit risk is one major threat to financial institutions \cite{buehler2008new,li2015aggregation,ma2018study,tan2018deep}, better credit risk prediction also improves the risk management capacity of banks and financial technology companies.

Although credit scores, such as FICO Score, have been widely used as mainstream risk indicators by many financial institutions, data-driven methods have recently shown their great potential and superior practical performances~\cite{xu2021loan}. 
Deep learning (DL), the dominating modeling technique in various domains such as computer vision, natural language processing, and recommendation system, has been a promising and increasingly popular tool considered to tackle financial problems. Recent attempts include market prediction \cite{ding2015deep,minh2018deep}, stock trading \cite{sezer2017deep} and exchange rate prediction \cite{shen2015forecasting}. Despite the recent trend of using deep models, non-DL methods, such as XGBoost and logistic regression, remain the most effective techniques so far for credit risk prediction in the financial industry.
Many existing studies have shown that neural network models lead to similar or even worse performances than non-DL methods \cite{fu2017combination,kvamme2018predicting,varmedja2019credit,li2020heterogeneous,moscato2021benchmark}.

Credit risk prediction can be formulated as a binary classification problem, where the goal is to learn a function $f_\theta:\mathcal{X}\to[0,1]$ to map the credit information $x\in\mathcal{X}$ of an applicant to a risk score $y\in[0,1]$ that represents the probability of default. 

Despite such a simple problem formulation, credit risk prediction can be particularly challenging. Existing deep-learning-based solutions mainly focus on e-commerce consumer data~\cite{liang2021credit}, which typically include dense features and highly frequent user activities, such as clicks and payments, on e-commerce platforms. However, these fine-grained data are not commonly available to financial institutions. Specifically, in our application, we adopt the official credit reports provided by the Credit Reference Center (CRC) of the People’s Bank of China. These financial data are of much lower quality, i.e., containing much higher dimensions (over 4k) with a large portion of missing entries and extreme values, due to low-frequency credit records. 
End-to-end training neural networks on these data can be substantially more challenging and brittle~\cite{poole2016exponential,borisov2021deep}.
Therefore, to the best of our knowledge, most financial institutions (e.g., banks) still adopt non-DL-based methods.

In this work, we present a successful industrial case study by developing an effective deep learning framework, \emph{DeRisk},  which outperforms our production decision-tree-based system, on real-world financial data. Our \emph{DeRisk} framework consists of three major stages including data pre-processing, separate training of non-sequential and sequential models, and joint fine-tuning. We also design a collection of practical techniques to stabilize deep neural network training under the aforementioned challenges. Specifically for the low-quality real-world financial data, we observe that a multi-stage process with feature selection and DL-specific engineering processing can be critical to the overall success of our framework.

\textbf{Main contributions.} (1) We develop a \textbf{comprehensive workflow} that \emph{considers all the model training aspects} for risk prediction. (2) We implement \emph{DeRisk}, the \emph{first} deep risk prediction model that \textbf{outperforms statistical learning approaches} on \emph{real-world financial data}. (3) We conduct \textbf{extensive ablation studies} on the effect of different technical components of \emph{DeRisk}, which \emph{provides useful insights and  practical suggestions} for the research community and relevant practitioners.

\section{Related Work}
There have been extensive studies using machine learning techniques for credit risk prediction, including linear regression \cite{puro2010borrower,guo2016instance}, SVM \cite{jadhav2018information,kim2019ensemble}, decision tree based methods like Random Forest (RF) \cite{malekipirbazari2015risk,varmedja2019credit,xu2021loan} or Gradient Boost Decision Tree (GBDT) \cite{xia2017boosted,he2018novel}, deep learning \cite{byanjankar2015predicting,kvamme2018predicting,yang2018deepcredit,yotsawat2021novel}, or an ensemble of them \cite{fu2017combination,li2020heterogeneous}. Most of these works use data with non-sequential features. Although deep learning is applied, empirical results find that XGBoost or other GBDT approaches usually outperforms deep learning \cite{fu2017combination,kvamme2018predicting,varmedja2019credit,xu2021loan}. 

On the other hand, deep learning has shown its superiority beyond tabular data through the flexibility of deep neural networks. Convolutional Neural Network (CNN) \cite{kvamme2018predicting}, Long Short-Term Memory (LSTM) \cite{yang2018deepcredit} and Graph Neural Network (GNN) \cite{wang2021temporal} are adopted for sequential data or graph data since other machine learning techniques like GBDT fail to properly model non-tabular data. According to \cite{liang2021credit}, deep learning outperforms conventional methods on multimodal e-commerce data for credit risk prediction.

Many data challenges in financial applications are also common in other machine learning fields. 
(1) For high-dimensional data, many feature selection methods have been proposed, including 
filter methods \cite{gu2011generalized}, wrapper methods \cite{yamada2014high} and embedded methods \cite{feng2017sparse}. 
Many risk prediction works have adopted feature selection for better performance \cite{xia2017boosted,ha2019improving,li2020heterogeneous} or interpretability \cite{ma2018study,xu2021loan}. 
(2) Handling multiple data formats and feature types is related to the field of deep learning for tabular data~\cite{gorishniy2021revisiting,borisov2021deep}.
There are typical three popular deep neural network architectures for tabular data \cite{klambauer2017self,huang2020tabtransformer,arik2021tabnet}, including Multi-Layer Perception (MLP), Residual Network (ResNet) \cite{he2016deep} and Transformer \cite{vaswani2017attention}. Similar to the financial domain, it is also reported that deep models are not universally superior to GBDT models \cite{gorishniy2021revisiting} on tabular data.
(3) For the out-of-time distribution shift issue, it is common to split training and test data according to the temporal order~\cite{kvamme2018predicting,jiang2021financial}.
(4) Furthermore, data imbalance is also a long-standing problem in machine learning research. Among the popular over-sampling and under-sampling strategies \cite{he2018novel,bastani2019wide,mahbobi2021credit}, Synthetic Minority Over-sampling Technique (SMOTE) \cite{chawla2002smote} is a widespread technique for synthetic minority data, which is also reported be effective for credit risk prediction \cite{bastani2019wide}. Generative adversarial networks can also be used to generate additional minority data \cite{mariani2018bagan} and this method can be applied to financial data \cite{liu2020alike} for risk prediction. However, these methods are limited to non-sequential data generation, while our financial data has multiple formats. Class-balanced loss is another method to make the model attend more to the minority samples \cite{lin2017focal,xia2017cost,cui2019class,ren2022balanced}. Comparative experiments \cite{kaur2019systematic,moscato2021benchmark} show that all strategies have their pros and cons. In our work, we use a class-balanced loss to mitigate the problem of data imbalance, and different strategies are used for non-sequential data and sequential data thanks to their great difference in data dimension. 
\section{Preliminary}
\label{sec:dataset}

In this section, we first present the problem statement for the credit risk prediction task, and then introduce the credit information and labels used in the task. \\
\subsection{Task Formulation}
The credit risk prediction task aims to decide whether a loan can be granted to the applicant according to his/her credit information. To be more specific, the risk prediction model needs to learn a function $f_\theta:\mathcal{X}\to[0,1]$, which takes the credit information $x\in\mathcal{X}$ of an applicant as input and produces a risk score $y\in[0,1]$ that represents the probability of delinquent on the applicant's payments. \\
\subsection{Multi-format Credit Information.}
In this work, we adopt the credit information in the credit report data that is generally available in financial institutions. The credit report data of an applicant consists of two parts: non-sequential features and sequential features. Specifically, the non-sequential part usually contains thousands of stable profiles of the applicant, including age, marital status, industry, property status, etc. We remark that the non-sequential data of a credit report can be extremely high-dimensional and sparse, which requires further processing to successfully train deep neural network models. The sequential part contains dozens of features and consists of three components of the applicant's financial behavior organized by time: (1) applicant's past loan information (\textbf{loan}), including the date of loan issuing, type of lending institution, loan amount, etc.; (2) the records that applicant's credit report was inquired in the past (\textbf{inquiry}), including inquiry time, inquiry institutions, inquiry reasons, etc.; (3) applicant's credit card information (\textbf{card}), including card application date, credit card type, currency, etc. Note that the number of sequential features is much smaller than non-sequential features. \\
\subsection{Multiple Labels and Imbalanced Data } 
Loan repayments naturally generate multiple labels because of installment (e.g., the first or the second month to pay back) and different degrees of delinquency (e.g., one-week or one-month delay). These labels are roughly categorized into short-term labels (e.g., the first/second/third installment is more than 30 days overdue) and long-term labels (e.g., any installment in recent 12 months is more than 5/15/30 days overdue). Due to the general priority of short-term benefits and the convenience of subsequent collection, financial institutions typically use short-term labels for evaluation. However, directly using this short-term evaluation label as the training label can be suboptimal. The choice of training label needs careful consideration for the best practice. Note that all these labels are particularly imbalanced (10\% or even 1\% for minority samples) because applicants who pay on time are much more than applicants who are overdue. Therefore, different choices of labels may lead to drastically different model performances in practice, as shown in our ablation study in Section~\ref{analysis-label}.

\section{Methodology}
\subsection{Overall Pipeline}
The overall pipeline of our \emph{DeRisk} framework is shown in Figure \ref{fig:pipline}. Firstly, we apply careful data processing to turn noisy and irregular input features into a neatly structured format, which is indispensable for training deep networks. Secondly, to well utilize both sequential and non-sequential features, we design two main sub-models: a DNN model for processing non-sequential features and a Transformer-based model $\mathcal{M}_\text{S}$ for processing sequential features. We train them separately in the second stage. In the last stage, we fuse $\mathcal{M}_\text{NS}$ and $\mathcal{M}_\text{S}$ by concatenating the final hidden layers from both models and applying another linear head to give the final prediction score. We jointly fine-tune this whole model to get improved performance.

\subsection{Selection of Training Label}
\label{sec:method-label}
As we mentioned in Sec. \ref{sec:dataset}, there are multiple labels in risk prediction tasks that record an applicant's repayment behavior in different time periods. Among these labels, we choose a long-term label to train our model for two reasons. First, long-term labels are more balanced than short-term labels. Second, the data distribution (e.g., the ratio of negative and positive data) varies over time (see Appendix \ref{sec:vary}) because of economic changes and the continual improvement of our deployed model. 
The long-term label is less sensitive to these influences and is more stable because it summarizes an applicant's behavior in the last 12 months, conceptually performing a smoothing operator over the timeline. We believe this will make our model more generalizable and perform better on the out-of-time test set, though predicting long-term risk is inherently more difficult. 

\subsection{Data Pre-Processing}
The credit report data, especially the non-sequential data, is extremely complex and noisy, as it contains many missing values and outlier values. This low-quality input can make the learning process unstable and hurts the final performance. Therefore, proper data pre-preprocessing can be significantly beneficial for the optimization of DL models.

Both sequential and non-sequential features can be divided into three types: time features (i.e., features about time such as credit card issue date), real-value features (e.g., age, loan amount), and category features (e.g., industry, type of lending institution). For the time features, we always use a relative date difference to avoid the models memorizing input data according to the date. We also apply normalization for the numerical time features and real-value features, and discard minor classes in the category features. 

In addition, we adopt specific techniques for non-sequential features. We found that lots of non-sequential real-value features are useless noise and even harmful for training. Hence we adopt a commonly-used feature selection technique that utilizes XGBoost~\cite{chen2015xgboost} to select the most important 500 features among thousands of non-sequential real-value features and discard the others. Besides, most non-sequential features have many 0s and missing values (NAN) that naturally arise from the financial behaviors and data collection processes, which makes non-sequential data sparse, noisy, and problematic for DL training. These 0s and NANs are not necessarily meaningless, e.g., a NAN in ``The time of first application for a mortgage" may imply that this applicant has never applied for a mortgage. Besides, if we simply fill these entries with a constant $c$, it will influence those true entries close to $c$ and significantly influence the learned model. 
So, we treat these 0s and NANs carefully. For every category feature, we add a category $\langle \text{NAN} \rangle$, and for every real-value and time feature, besides replacing all NANs with 0s, we also create two indicators that directly tell whether a value is 0 and is NAN. With explicit indicators, DL models can therefore directly utilize the information implied by meaningful 0s and NANs and learn to ignore those 0s and NANs that are harmful to training. 

\begin{figure}[!t]
  \centering
  \includegraphics[width=0.4\textwidth]{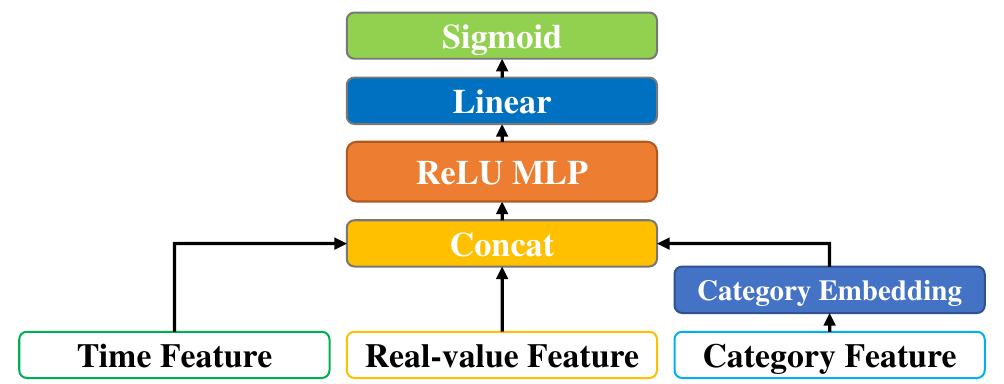}
  \caption{Non-sequential DNN model.}
  \label{fig:nonseq_model}
\end{figure}

\subsection{Modeling Non-sequential Features}

We adopt a simple but effective neural network for non-sequential features. 
The architecture is shown in Figure \ref{fig:nonseq_model}. Firstly it uses an embedding layer to convert category features into dense vectors and concatenate them with time and real-value features to the dense input $x_\text{dense}^\text{NS}\in \mathbb{R}^{m_1}$. Then $x_\text{dense}^\text{NS}$ is fed into a MLP (multi-layer perceptron) with ReLU activation function to get the non-sequential output hidden state $x_\text{final}^\text{NS}\in \mathbb{R}^{m_2}$. And the final prediction $\hat y^\text{NS}$ is computed as: $\hat y^\text{NS}=\sigma((\text{w}_\text{logit}^\text{NS})^Tx_\text{final}^\text{NS}+\text{b}_\text{logit}^\text{NS})$, where $\text{w}_\text{logit}^\text{NS}\in \mathbb{R}^{m_2}$ and $\text{b}_\text{logit}^\text{NS}$ are the weight vector and bias for the logit, respectively, $z^\text{NS}=(\text{w}_\text{logit}^\text{NS})^Tx_\text{final}^\text{NS}+\text{b}_\text{logit}^\text{NS}$ is the logit, and $\sigma(x)=1/(1+\exp(-x))$ is sigmoid activation. 

\begin{figure}[!t]
  \centering
  \includegraphics[width=0.4\textwidth]{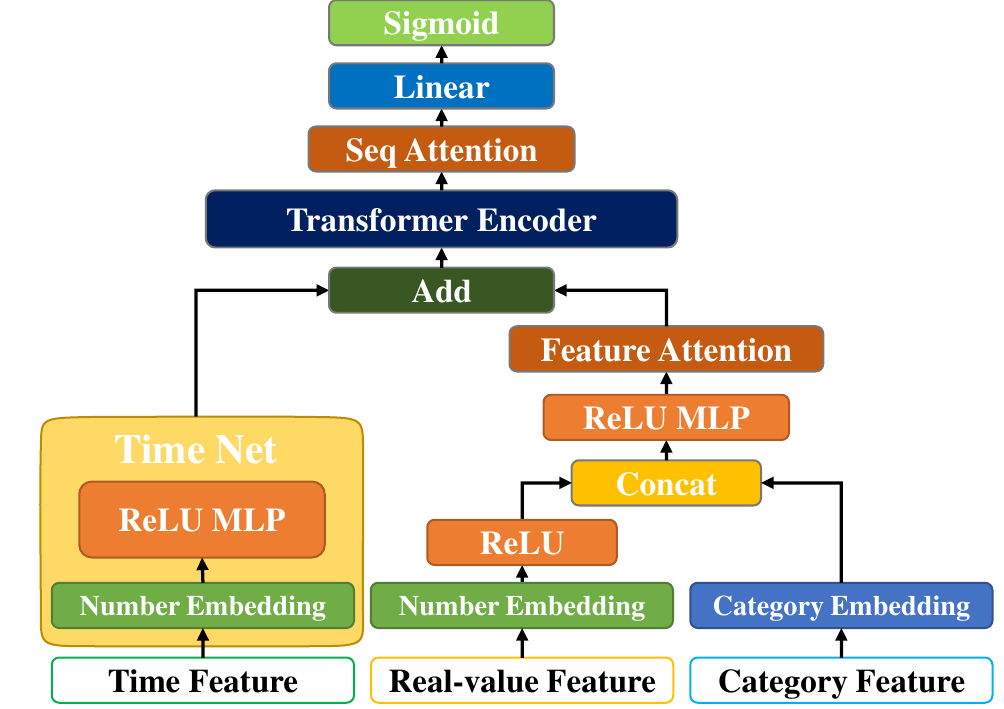}
  \caption{Sequential Transformer-based model.}
  \label{fig:seq_model}
\end{figure}

\subsection{Modeling Sequential Features}

\subsubsection{Architecture}
We adopt a Transformer~\cite{vaswani2017attention}-based model for its strong modeling capacity. 
The architecture is shown in Figure \ref{fig:seq_model}. Three such models, $\mathcal{M}_\text{card}$,  $\mathcal{M}_\text{inquiry}$, and $\mathcal{M}_\text{loan}$, are used for card, inquiry, and loan features, respectively. Suppose the sequence length is $l$ and the embedding size is $e$. Firstly a time net will convert the time feature into time embedding $E_t\in \mathbb{R}^{l\times e}$, which plays a role of position embedding, and attention is used to merge different feature embeddings into one, i.e., $E_f\in \mathbb{R}^{l\times e}$. Then a Transformer encoder will encode the sequential embeddings $E=E_t + E_f$ into hidden feature $x_h\in \mathbb{R}^{l\times e}$, which will be pooled by another attention into output feature $x_\text{final}^\text{*}\in \mathbb{R}^e$, where $*$ refer to card, inquiry or loan. We concatenate $x_\text{final}^\text{card}$,  $x_\text{final}^\text{inquiry}$, and $x_\text{final}^\text{loan}$ to obtain $x_\text{final}^\text{S}\in \mathbb{R}^{3e}$. At last, similar to non-sequential case, we have logit $z^\text{S}=(\text{w}_\text{logit}^\text{S})^Tx_\text{final}^\text{S}+\text{b}_\text{logit}^\text{S}$ and final predication $\hat y^\text{S}=\sigma(z^\text{S})$. To improve the generalization ability of the sequential model, we share the time net and Transformer encoder among $\mathcal{M}_\text{card}$,  $\mathcal{M}_\text{inquiry}$, and $\mathcal{M}_\text{loan}$. 

\begin{figure*}[!t]
  \centering
  \includegraphics[width=0.8\textwidth]{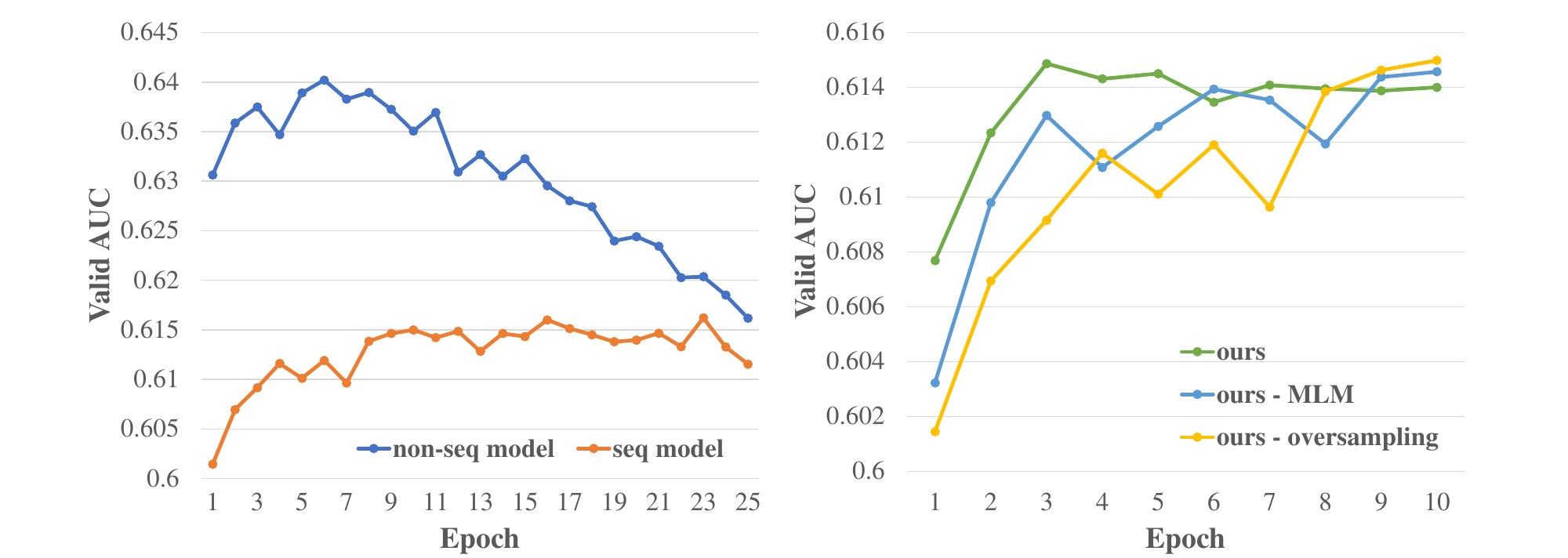}
  \caption{Left: valid AUC of non-sequential and sequential models with training epoch. Right: change of valid AUC of the sequential model with and without MLM pre-training or oversampling. }
  \label{fig:mlm}
\end{figure*}

\subsubsection{Mask Language Model Pre-training}
During training, we found that optimization of the sequential model is much harder than the non-sequential model (the left part of Figure \ref{fig:mlm}) due to the scarcity of sequential features compared with non-sequential features. To ease the training of the sequential model, we adopt mask language model (MLM) pre-training as BERT~\cite{devlin2018bert} to make the model first learn informative and general features from sequential data. We randomly mask the input sequential features, where $80$\% of masked value are replaced with token $\langle \text{MASK} \rangle$ (for category features) or 0 (for time and real-value features), 10\% are replaced with a random value, and 10\% remain unchanged. The three output hidden features of Transformer encoder, i.e., $x_h^\text{card}$,  $x_h^\text{inquiry}$, and $x_h^\text{loan}$, will be input into different classification heads to predict different type of origin value at the masked position. After pre-training, we fine-tune $\mathcal{M}_\text{NS}$ on the downstream classification task. 

\subsection{Weighted BCE Loss} 
We also adopt weighted BCE loss to deal with data imbalance. Firstly, BCE (binary cross entropy) loss function is commonly used in binary classification tasks: 
$$\text{BCE}=-\frac{1}{n}\sum_{i=1}^n[y_i\log(\hat y_i)+(1-y_i)\log(1-\hat y_i)].$$ However, when negative samples are much more than positive samples, naive BCE loss will induce the model to output $\hat y_i=0$. To avoid this, we can give more weight to positive samples by using weighted BCE loss: 
$$\text{WBCE}=-\frac{1}{|\mathcal{D}^-|}\sum_{i\in \mathcal{D}^-}\log(1-\hat y_i)-\frac{1}{|\mathcal{D}^+|}\sum_{i\in \mathcal{D}^+}\log(\hat y_i),$$
where $\mathcal{D}^+=\{i:y_i=1\}$ is the set of positive samples and $\mathcal{D}^-=\{i:y_i=0\}$ is the set of negative samples. Another implementation of the above-mentioned weighted loss is oversampling, i.e., adjusting the ratio $|\mathcal{D}^-|:|\mathcal{D}^+|$ to $1 : 1$ by re-sampling positive samples. 

We use oversampling on the sequential model and use normal weighted BCE loss on the non-sequential model and joint fine-tuning stage. This is because the optimization of the non-sequential model is much harder and slower than that of the non-sequential model due to the small number of non-sequential features, while oversampling enables the model to see rare samples multiple times in one epoch and thus accelerates optimization. On the other hand, the number of non-sequential features is large and the optimization of the non-sequential model is already fast enough, oversampling may lead to overfitting on the minority samples instead.

\subsection{Separate Training \& Joint Fine-tuning}
To fuse the sequential and non-sequential features, we use a concatenation layer (\textbf{Concat Net}) on the top of them to concatenate their output hidden states and to predict the final score, i.e., $\hat y=\sigma((\text{w}_\text{logit})^Tx_\text{final}+\text{b}_\text{logit})$, where $x_\text{final}^\text{NS}=[x_\text{final}^\text{NS}, x_\text{final}^\text{S}]$. Note that the hardness of optimization non-sequential and sequential models is different, so if we train them with the concatenation layer together from scratch, the overall model will totally rely on the non-sequential outputs, which are easier to train on, while ignoring the output of the sequential model. To avoid this and to utilize the sequential features better, we adopt a two-stage training strategy: separately train sequential and non-sequential models first and then jointly fine-tune them with the Concat Net. 

\section{Experiment Setup}
\label{sec:experiment}

\begin{table}[!t]
\centering
\begin{tabular}{l|ccc}
\toprule
\textbf{Data}   & Time & Real-value & Category\\ \hline
 card   & 1 & 2  & 5   \\
                            inquiry  & 1 & 0 & 2       \\
                            loan &  1 & 4 & 5   \\\hline
non-sequential  & 13  & 4098 & 9  \\\bottomrule       
\end{tabular}
\caption{Number of time, real-value, and category features in each sample of sequential and non-sequential data. The card, inquiry, and loan sequences for each user are clipped with lengths 32, 64, and 128, respectively.}
\label{table:number}
\end{table}


\subsection{Notation} 

We mainly use a long-term label $\mathcal Y_\text{long}$ for training and a short-term label $\mathcal Y_\text{short}^\text{eval}$ for evaluation. $\mathcal Y_\text{short}^\text{other1},\mathcal Y_\text{short}^\text{other2},\mathcal Y_\text{short}^\text{other3},$ There are also three other short-term labels used in our experiments. The description of these labels are in Sec~\ref{sec:label-notation}.

\subsection{Dataset Statistics}

\label{sec:label_notatoin}

We sample 582,996 Yanqianguan Users and use their credit report data and repayment behavior from August 2020 to July 2021 as the dataset. To simulate the out-of-time prediction in real business scenarios, we take the 430,865 data pieces from August 2020 to May 2021 as the training set and 152,131 data pieces from June 2021 to July 2021 as the test set. The ratio of negative and positive samples is about $50:1$ according to the short-term label used for evaluation and is about $10:1$ according to the long-term label used for training.\footnote{We keep the exact ratio numbers confidential due to commercial and security concerns.} For sequential data, we set the maximum sequence length of card, query, and loan data to be 32, 64, and 128, respectively, according to the distribution of data length. Only the latest data will be included for training and evaluation. Some statistics are summarized in Table \ref{table:number}.

Note that all above data are definitely authorized by the customers since they hope to apply for loan in our platform and they should provide the access to their credit report. We also anonymized the names of people and organizations on credit reports to protect customers' privacy.


\subsection{Evaluation Metric}
The metric commonly used to evaluate credit risk prediction models $\mathcal{M}$ is AUC (Area Under the ROC Curve) score.
We remark that this is a challenging task and an increment of 0.01 in AUC can be significant in performance as this results in a roughly 5\% decrement of real-world bad debts.

\section{Main Results}
\begin{table}[!t]
  \centering
    \begin{tabular}{lccc}
    \toprule
    \textbf{Model}   & \textbf{$\mathcal Y_\text{short}^\text{eval}$} & \textbf{$\mathcal Y_\text{short}^\text{other2}$} & \textbf{$\mathcal Y_\text{short}^\text{other3}$} \\[1ex]\hline
    \multicolumn{4}{l}{\textit{non-seq model over non-seq data only}}                                                                                                              \\\hline
    XGBoost               & 0.6418             & 0.6282             & 0.6187             \\
    DeepFM                                & 0.5700             & 0.5508             & 0.5478             \\
    SDCN                                & 0.6450             & 0.6319             & 0.6236             \\
    PDCN                                & 0.6483             & 0.6343             & \textbf{0.6254}    \\
    AutoInt                                & 0.6454             & 0.6325             & 0.6238    \\
    DNN                                 & \textbf{0.6499}    & \textbf{0.6349}    & \textbf{0.6254}    \\\hline
    \multicolumn{4}{l}{\textit{seq model over sqe data only}}                                                                                                                  \\\hline
    Pooled MLP                            & 0.5996            & 0.5821             & 0.5749             \\
    LSTM                               & 0.6108             & 0.5936             & 0.5859             \\
    Transformer                           & 0.6132             & 0.5941             & 0.5871             \\
    Transformer+MLM                     & \textbf{0.6156}    & \textbf{0.5971}    & \textbf{0.5885}    \\\hline
    \multicolumn{4}{l}{\textit{joint model over the entire data}}                                                                                                                \\\hline
    Add-Attn Net                        & 0.6504             & 0.6369             & 0.6285             \\
    Mul-Attn Net                        & 0.6520             & 0.6377             & 0.6278             \\
    \emph{DeRisk}(ours)                          & \textbf{0.6546}    & \textbf{0.6398}    & \textbf{0.6297}    \\
    \bottomrule
    \end{tabular}
    \caption{All models are evaluated by AUC scores on three different short-term labels.} 
    \label{table:main}
\end{table}

\subsection{Baselines}
\label{sec:exp-baselines}
For non-sequential model, the baselines include (1) current popular traditional ML model \textbf{XGBoost}~\cite{chen2015xgboost} (main baseline) and several more complicated deep models including  (2) \textbf{DeepFM}~\cite{guo2017deepfm}: the final score is $y^\text{NS}=\sigma(z^\text{NS}_\text{DNN}+z^\text{NS}_\text{FM})$, where  $z^\text{NS}_\text{DNN}$ is the logit of DNN and $z^\text{NS}_\text{FM}$ is the logit gotten by a FM (factorization machine~\cite{rendle2010factorization}) layer. 
    (3) \textbf{DCNv2}~\cite{wang2021dcn}: use cross-network (multiple cross layers) to obtain high-order cross feature.  A DNN can be stacked on top of the cross-network (SDCN); we could also place them in parallel (PDCN).
    (4) \textbf{AutoInt}~\cite{song2019autoint}: use a multi-head self-attention to learn interacted features.

For the sequential model, our baselines are \textbf{pooled MLP} (which uses a pooling layer to average hidden states of different times that are individually produced by the MLP) and \textbf{LSTM}~\cite{hochreiter1997long}.

For the final module that fuses the output of the hidden state by the non-sequential model and sequential model, we compare our simple Concat Net with an additive attention layer (\textbf{Add-Attn Net}) and a multiplicative attention layer (\textbf{Mul-Attn Net}) that use $x_\text{final}^{NS}$ as a query vector to pool output hidden feature of Transformer Encoder $x_h$ by additive and multiplicative attention, respectively.

\subsection{Evaluation and Analysis}

Since our dataset has multiple formats, we first test separated models for single-format data modeling. For non-sequential data, we compare the DNN module in \textit{DeRisk} with XGboost, a widely-used decision-tree model in our production system. We aim to show whether our \textit{DeRisk} system and techniques can make its DNN module outperform other non-DL methods on real-world financial data. Other popular models in recommendation systems like DeepFM, DCN, and AutoInt are also tested as DL competitors. For sequential data, we consider different sequential models including Pooled MLP, LSTM, and Transformer for evaluation. Our \textit{DeRisk} adopts Transformer and additionally adopts MLM-pretraining to accelerate training.

Finally, we consider joint models trained over the entire dataset with both formats by fusing the best non-sequential model, DNN, and the best sequential model MLM-pretrained Transfomer, to obtain joint models for the best evaluation results. With more data, the joint models outperform either separated models, but we also find different fusing techniques lead to different performances. We compare our Concat Net with two different attention-based methods.

Table \ref{table:main} summarizes the main results. 
All models are evaluated by three different labels to show consistent results. From the results we can see that: 
\begin{itemize}
    \item[(1)] Our non-sequential model DNN and sequential model \\ MLM+Transformer outperform all baselines, respectively. Specifically, compared with current popular XGBoost model, our DNN model $\mathcal{M}_{\textrm{NS}}$ and best joint model \emph{DeRisk} (with Concat Net) improve $\mathcal Y_\text{short}^\text{eval}$ AUC score by 0.0081 and 0.0128, respectively.  
    \item[(2)] Joint fine-tuning of non-sequential and sequential models can achieve better results than only using a single non-sequential or sequential model.
    \item[(3)] Complex models do not necessarily perform better: simplest DNN and Concat Net outperform other more complicated models. This indicates that the high-order features created by those additional networks such as FM and cross layers are not that helpful for the credit risk prediction task.

\end{itemize}

\section{Ablation Study}

In this section, we conduct a series of experiments to demonstrate the effect of each part of our \emph{DeRisk} framework. We mainly use $\mathcal Y_\text{short}^\text{eval}$ for evaluation since we find it shows a consistent result with other short-term labels as in Table \ref{table:main}. We test the effectiveness of different modules in our multi-stage process, including separate training \& joint fine-tuning, feature selection, indicator features, and MLM-pretraining. Many different techniques for data imbalance are also studied in this section. With our ablation studies, we also present best practices for training deep neural network models over real-world financial data.


\subsection{Effect of Multi-stage Training}

\begin{table}[tb]
\centering
\begin{tabular}{lc}
\toprule
\textbf{Change}  & \textbf{AUC}    \\\hline
No (ours) & \textbf{0.6546} \\
w/o Separate Training (end-to-end)  & 0.6487        \\
w/ Freeze Sub-models & 0.6512      \\\bottomrule   
\end{tabular}
\caption{$\mathcal Y_\text{short}^\text{eval}$ AUC scores with different training strategies.}
\label{table:joint}
\end{table}

Because the hardness of optimization on non-sequential data and sequential data is different as shown in Figure \ref{fig:mlm}, we first separately train $\mathcal{M}_\text{NS}$ and $\mathcal{M}_\text{S}$ and then joint fine-tune them. We also tried joint training them from scratch (end-to-end), or freezing $\mathcal{M}_\text{NS}$ and $\mathcal{M}_\text{S}$ and only tuning the concatenating layer during joint fine-tuning. The results are reported in Table \ref{table:joint}. We can see that separate training outperforms the other two training strategies. \\ \textbf{Suggestion\#1: It is beneficial to first perform separate training and then joint tuning for multi-format data. The additional tunable parameters introduced in the fine-tuning process should be sufficiently large for effective multi-format fusion. }

\subsection{Effect of Different Training Labels}
\label{analysis-label}

\begin{table}[tb]
\centering
\begin{tabular}{ccc}
\toprule
\textbf{Training Label} & \textbf{Test Label} & \textbf{AUC}    \\\hline
    \multicolumn{3}{l}{\textit{non-seq model}}                          \\\hline\\[-2.5ex]
$\mathcal Y_\text{long}$ (Ours) & $\mathcal Y_\text{short}^\text{eval}$ & \textbf{0.6499} \\[1.0ex]
 $\mathcal Y_\text{short}^\text{other1}$     & $\mathcal Y_\text{short}^\text{eval}$   & 0.6392          \\[1.0ex]
 $\mathcal Y_\text{short}^\text{eval}$      & $\mathcal Y_\text{short}^\text{eval}$  & 0.6363   \\[1.1ex]\hline
    \multicolumn{3}{l}{\textit{seq model}}                          \\\hline\\[-2.5ex]
$\mathcal Y_\text{long}$ (Ours)               & $\mathcal Y_\text{short}^\text{eval}$ & \textbf{0.6156} \\[1.1ex]
  $\mathcal Y_\text{short}^\text{other1}$     & $\mathcal Y_\text{short}^\text{eval}$ & 0.6113          \\[1.1ex]
 $\mathcal Y_\text{short}^\text{eval}$   & $\mathcal Y_\text{short}^\text{eval}$ & 0.6105          \\[0.4ex]\bottomrule 
\end{tabular}
\caption{Experiment results of selecting different training labels on non-sequential and sequential models.}
\label{table:label}
\end{table}

We tried taking two short-term labels ($\mathcal Y_\text{short}^\text{other1}$ and $\mathcal Y_\text{short}^\text{eval}$) and a long-term label ($\mathcal Y_\text{long}$) as the training label, respectively. The results in Table \ref{table:label} demonstrate that the long-term label is the best choice for both non-sequential and sequential models, even when the model is evaluated on a short-term label.\\ \textbf{Suggestion\#2: It is better to choose a balanced and stable signal that measures the long-term behaviors as the training label.}

\subsection{Effect of Real-value Feature Selection}

\begin{table}[tb]
\centering
\begin{tabular}{lc}
\toprule
\textbf{Change}  & \textbf{AUC}    \\\hline
No (Ours)        & \textbf{0.6499} \\
$|\mathcal{F}_R|=4098$                  & 0.6415          \\
$|\mathcal{F}_R|=100$                    & 0.6390          \\
w/o Indicator               & 0.6426          \\
w/ BCE Loss                & 0.6454          \\
w/ Focal Loss                     & 0.6403          \\
w/ Oversample            & 0.6458    \\\bottomrule     
\end{tabular}
\caption{Analysis experiment results on non-sequential DNN model, where $|\mathcal{F}_R|$ is the number of selected features.}
\label{table:dnn}
\end{table}

To show the effect of selecting real-value features with XGBoost, we compare the following three cases: no selection, selecting 500 real-value features (Ours), and selecting 100 real-value features. The results in Table \ref{table:dnn} show that selecting 500 features performs the best. This indicates that (1) by selecting real-value features with XGBoost, we can drop useful fewer features and improve the performance. (2) dropping too many features would lead to worse predictions. \\\textbf{Suggestion\#3: It is important to perform feature selection before deep learning training. The dimension of selected features should be chosen carefully}.

\subsection{Effect of Indicator Features}
To show the effect of NAN and zero indicators, we compare the case with and without them. As shown in Table~\ref{table:dnn}, after removing indicators, the AUC score decreases by 0.0073. \\\textbf{Suggestion\#4: Some NANs and 0s can be meaningful and it is better to use indicator features rather than simply filling these missing values with a constant or discarding them.}

\subsection{Comparison of Different Loss Functions}
\label{Loss}
We compared the performance of using weighted BCE loss (Ours) with using naive BCE loss on the DNN model. In addition, we also tried Focal loss~\cite{lin2017focal} which is designed for the data imbalance case, but the result in Table \ref{table:dnn} shows that it is not helpful for our task and weight BCE achieves the best performance. \\
\textbf{Suggestion\#5: Adding more weight to rare positive samples is critical to prevent the model from biasing to the overwhelming negative outputs.} 

\subsection{Effect of Oversampling}

\begin{table}[!t]
\centering 
\begin{tabular}{lc}
\toprule
\textbf{Change} & \textbf{AUC} \\\hline
No (Ours)          & \textbf{0.6156} \\
w/o MLM Pre-training         & 0.6132          \\
w/o Oversampling  & 0.6153    \\\bottomrule     
\end{tabular}
\caption{Analysis experiment results on sequential Transformer-based model.}
\label{table:transfo}
\end{table}

We compared the cases with and without oversampling on both the non-sequential model and sequential model to demonstrate the effect of oversampling. We can see from Table \ref{table:transfo} and the right of Figure \ref{fig:mlm} that for sequential model, oversampling  (1) improves AUC. (2) accelerates optimization. By enabling the model to see rare positive samples more times in each epoch, oversampling reduces the training difficulty of the sequential model. On the other hand,  oversampling also makes the non-sequential model, the one easier to optimize, overfits more quickly on the training data and thus cannot achieve good performance as shown in Table \ref{table:dnn}. In practice, DNN with oversampling usually overfits after the first epoch.  \\
\textbf{Suggestion\#6: Oversampling makes optimization of the sequential model easier and improves performance. And considering the difference between non-sequential data and sequential data, each separated model should be optimized with different sampling strategies.}

\subsection{Effect of MLM Pre-training of Sequential Model}
From Table \ref{table:transfo} and the right of Figure \ref{fig:mlm} that MLM pre-training of the sequential model  (1) improves performance. (2) accelerates optimization. This indicates that the pre-trained model has learned some knowledge of sequential data that are useful for the risk prediction task. \\ \textbf{Suggestion\#7: MLM pre-training benefits the optimization of the sequential model on credit risk prediction.}  

\section{Conclusion}
In this work, we proposed an effective deep learning framework, \emph{DeRisk}, which utilizes both sequential and non-sequential features for credit risk prediction. We apply careful data pre-processing to obtain clean and useful data for deep models, use MLM to pre-train the sequential model, adopt weighted BCE loss and oversampling to deal with the data imbalance problem, and select generalizable and stable training labels for better performance. 
The overall performance of \emph{DeRisk} largely outperforms existing approaches on real-world financial data. 
We remark that it is unnecessary that a more complicated network always performs better. In our analysis, every components of the training framework including data pre-processing and a carefully designed optimization process are all critical to make deep learning models perform well on a real-world financial application.
We hope our framework and analysis can bring insights for a wide range of important commercial applications and inspire future research on developing more powerful deep learning tools for real-world industrial data. 


\bibliography{references}

\appendix

\section{Appendix}

\subsection{Training Details}
\label{sec:hyper}
For our model and all the deep-learning baselines, we use Adam~\cite{kingma2014adam} optimizer with learning rate $5\times 10^{-4}$ and weight decay $1\times 10^{-4}$. We set the batch size to $1,000$. For non-sequential model DNN, we set the embedding size to 16, use three-layer MLP, and set the hidden size to 1028, 256, and 128, respectively. For sequential models, we use a one-layer Transformer encoder, set the embedding size to 128, the number of heads to 8, and the dropout probability to be 0.1. We adopt a 5-fold cross-validation on the training set and evaluate the ensembled model on the test set.

Both sequential and non-sequential features are composed of time features (i.e., features about time such as date), real-valued features, and category features.  For every time feature in date format, we subtract it by the date at which the credit report is used for prediction. That is, the time feature indicates the number of days between when the financial activity happens and when the credit report is called. Then for every time and real-value feature, we do zero-mean and one-std normalization and clip all values into $[-4, 4]$ to make the distribution easier for DL models to learn. For every category feature, we merge all the categories outside the top 30 into one category $\langle \text{UNK} \rangle$.

We utilize XGBoost~\cite{chen2015xgboost} to select the most important 500 features of the non-sequential real-value features and discard the rest of them. We simply train an XGBoost model on the same task of risk prediction. After that, we choose 500 features with the highest feature importance value to feed the non-sequential DL model. For every category feature, we add a category $\langle \text{NAN} \rangle$, and for every real-value and time feature, besides replacing all NANs with 0s, we also create two indicators $[x=0]$ and $[x=NAN]$. Therefore, for every real-value and time feature, there will be three corresponding features after this process. Thus, the 500 features we selected above become 1500 features.

\subsection{Label Notation}
\label{sec:label-notation}

The dataset mainly contains two types of labels: 1) short-term label i$x$label$y$, which means the user fails to pay back $y$ days after the $x$th-month's repayment deadline; 2) long-term label overdue$y$, which means the user has at least one $y$-day overdue behavior in the last year.

In the following experimental parts we mainly use the following labels: $\mathcal Y_\text{long}$, the long-term label overdue15; $\mathcal Y_\text{short}^\text{eval}$, the main short-term label i1label30 used for evaluation; $\mathcal Y_\text{short}^\text{other1},\mathcal Y_\text{short}^\text{other2},\mathcal Y_\text{short}^\text{other3},$ denoting another three short-term labels i1label15, i2label30, i3label30, for training and evaluation.

\subsection{Dataset Analysis}

\begin{figure}[!t]
  \centering
  \includegraphics[width=0.5\textwidth]{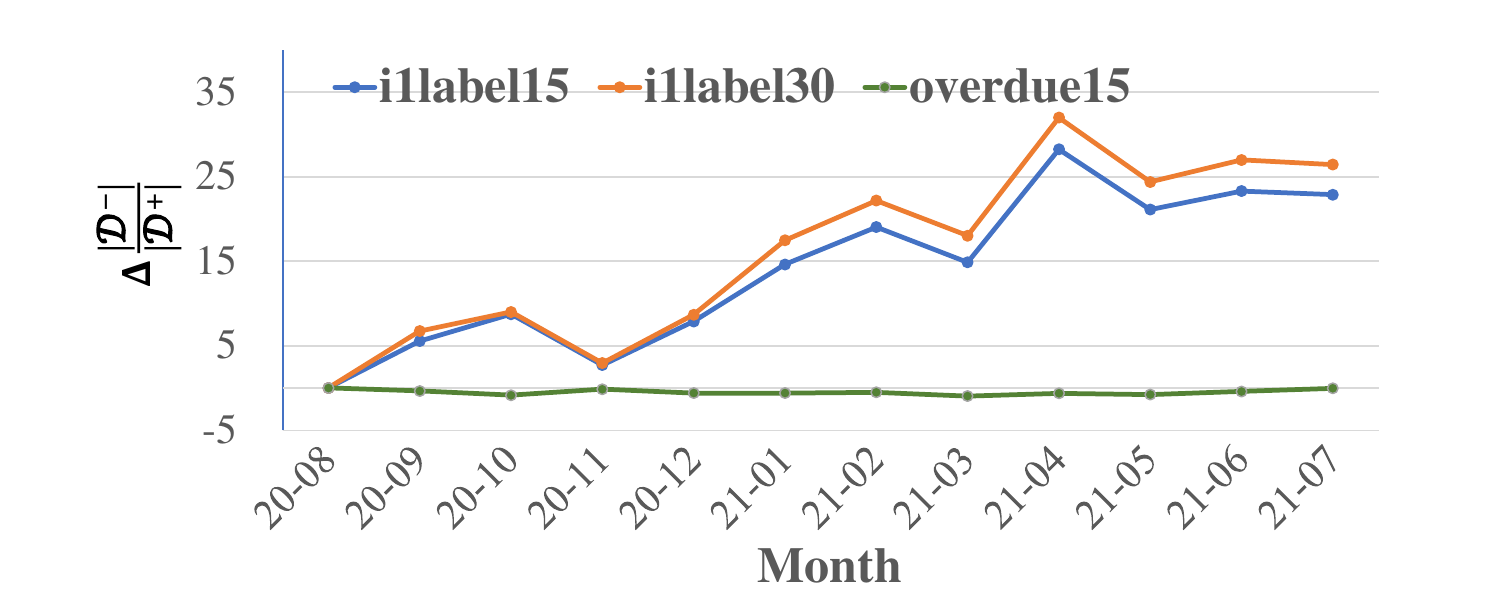}
  \caption{The change of imbalance ratio $\Delta \frac{|\mathcal{D}^-|}{|\mathcal{D}^+|}$ from August 2020 to July 2021. Compared with i1label15 and i1label30, the ratio of overdue15 is more stable.}
  \label{fig:imbalance}
\end{figure}

\label{sec:vary}
We show in  Figure~\ref{fig:imbalance} that the input data distribution, i.e., the ratio of negative and positive data, varies over time. Besides the changes of the economic environment, the data distribution changes also because the consumers are first filtered by a basic decision model in practice, which keeps being optimized over time. As a result of a better filtering process, fewer applicants default and the data becomes more imbalanced. (e.g., see Jan-2021 and May-2021 for i1label15 and i1label30).
 Empirically, compared to the short-term label, we notice that the long-term label overdue15 is less sensitive to economic environment influence and optimization of the basic decision model. It is more stable because it summarizes a customer's behavior in the last 12 months, which is conceptually performing a smoothing operator over the timeline. In addition, the prediction of long-term risk is more difficult and thus is less affected by the basic decision model.

\end{document}